\pdfoutput=1

\documentclass[11pt]{article}
\usepackage[]{acl2022}
\usepackage{times}
\usepackage{latexsym}
\usepackage[T1]{fontenc}
\usepackage[utf8]{inputenc}
\usepackage{microtype}

\usepackage{graphics}
\usepackage{graphicx}

\usepackage{amsmath}
\usepackage{amsfonts,amssymb}
\usepackage[ruled, linesnumbered]{algorithm2e}
\usepackage{amsmath}
\usepackage{booktabs}
\usepackage{multirow}
\usepackage{color}
\usepackage{makecell}
\usepackage{caption}
\usepackage{subcaption}
\usepackage{cleveref}
\usepackage{titlesec}
\usepackage{dsfont}

\DeclareMathOperator*{\argmin}{argmin}

\usepackage{microtype}

\usepackage{graphicx,txfonts}

\newcommand{\heart}{\ensuremath\varheartsuit}

\title{Faster Nearest Neighbor Machine Translation}
\author{Shuhe Wang$^{\heart\spadesuit}$, Jiwei Li$^{\clubsuit\spadesuit}$, Yuxian Meng$^{\spadesuit}$, Rongbin Ouyang$^{\heart}$\\ 
\bf Guoyin Wang$^{\blacktriangledown}$, Xiaoya Li$^{\spadesuit}$, Tianwei Zhang$^{\blacklozenge}$, Shi Zong$^{\blacktriangle}$ \\
$^\heart$Peking University, $^\clubsuit$Zhejiang University,$^\spadesuit$Shannon.AI \\
$^\blacktriangledown$Amazon Alexa AI, 
$^\blacklozenge$Nanyang Technological University,
$^\blacktriangle$Nanjing University\\
\{shuhe\_wang, jiwei\_li, yuxian\_meng, xiaoya\_li\}@shannonai.com, ouyang@pku.edu.cn\\
guoyiwan@amazon.com, tianwei.zhang@ntu.edu.sg, szong@nju.edu.cn}

\begin{document}

\maketitle

\begin{abstract}

$k$NN based neural machine translation ($k$NN-MT) has achieved state-of-the-art results in a variety of MT tasks. One significant shortcoming of $k$NN-MT lies in its inefficiency in identifying the $k$ nearest neighbors of the query representation 
from the entire datastore, which is prohibitively time-intensive when the datastore size is large.

In this work, we propose \textbf{Faster $k$NN-MT}  to address this issue. 
The core idea of Faster $k$NN-MT is to use a hierarchical clustering strategy to approximate the distance between the query and a data point in the datastore, which is decomposed into two parts: 
 the distance between the query and the center of the cluster that 
the data point belongs to, and the distance between the data point and the cluster center. 
We propose practical ways to compute these two parts in a significantly faster manner.
Through extensive experiments on different MT benchmarks, we show that \textbf{Faster $k$NN-MT} is 
faster than Fast $k$NN-MT \citep{meng2021fast} and only slightly (1.2 times) slower than its 
vanilla counterpart, while preserving model performance as $k$NN-MT.
Faster $k$NN-MT enables the deployment of  
 $k$NN-MT models on real-world MT services. 
\end{abstract}

\section{Introduction}
Recent years have witnessed the significant performance boost introduced by neural machine translation models \cite{sutskever2014sequence, cho2014learning,bahdanau2014neural, luong2015effective}. 
The recently proposed $k$NN based neural machine translation ($k$NN-MT) \citep{khandelwal2020nearest} has achieved state-of-the-art results across a wide variety of machine translation setups and datasets.
The core idea behind $k$NN-MT is that at each decoding step, the model is required to incorporate the target tokens with $k$ nearest translation contexts in a large constructed datastore. In short, $k$NN-MT refers to target tokens that come after similar translation contexts in the constructed datastore, leading to signficiant performance boost. 

One significant shortcoming of $k$NN-MT lies in its inefficiency in identifying the $k$ nearest neighbors from the whole target training tokens, which is prohibitively slow when the datastore is large. To tackle this issue, \citet{meng2021fast} proposed Fast $k$NN-MT. Fast $k$NN-MT evades the necessity of iterating over the entire datastore 
for the $K$NN search
by 
first
building smaller datastores for source tokens of a source sentence:
 for each source token, its datastore is limited to  
 reference tokens of the same token type, rather than the entire  corpus. 
 The concatenation of 
  the datastores for all source tokens are concatenated
  and mapped to corresponding target tokens, 
  forming the final datastore at the decoding step. 
 Fast $k$NN-MT is two-order faster than $k$NN-MT.
However, Fast $k$NN-MT needs to retrieve $k$ nearest neighbors of each query source token from all tokens of the same token type in the training set.
This can be still time-consuming when the current source reference token is a high-frequency word (e.g., “is”, “the”) and its corresponding
token-specific datastore is large. 
Additionally, the size of the datastore on the target side is propotional to the source length, making the model slow for long source inputs. 


In this paper, we propose Faster $k$NN-MT to address the aforementioned issues. 
The core idea of Faster $k$NN-MT is 
that we propose a novel 
 hierarchical clustering strategy to approximate the distance between the query and a data point in the datastore,
which is decomposed into two parts: 
(1) the distance between the query and the center of the cluster that 
the data point belongs to, and (2) the  distance between
the data point and the cluster centroid. 
The proposed strategy is both every effective in time and space. 
For (1), 
the computational complexity is low since the number of clusters is 
significantly smaller than the size of the datastore;
for (2),  distances 
between a cluster centroid and all constituent data points of that cluster can be computed in advanced and cached,  making (b) also fast.
Faster $k$NN-MT is also effective in space since 
it requires much smaller datastores than both Fast $k$NN-MT and vanilla $k$NN-MT. This makes it 
feasible to run the inference model with
a larger batch-size, which leads to an additional speedup.

Extensive experiments show that Faster $k$NN-MT
is only 1.2 times slower than standard MT model while preserving model performance.
Faster $k$NN-MT makes it feasible 
to deploy $k$NN-MT models on real-world MT services. 

The rest of this paper is organized as follows: 
we describe the background of $k$NN-MT and Fast $k$NN-MT in \Cref{sec:background}. The proposed Faster $k$NN-MT is detailed in \Cref{sec:our_method}. Experimental results are presented in \Cref{sec:exp}.
We briefly go through the related work in \Cref{sec:related_work}, followed by a brief conclusion in \Cref{sec:conclusion}. 

\section{Background}
\label{sec:background}



\subsection{$k$NN-MT}
\label{sec:knn-mt}

\paragraph{General MT.}
A general MT model translates a given input sentence $x=\{x_{1},...,x_{n}\}$ to a target sentence $y=\{y_{1},...,y_{m}\}$, where $n$ and $m$ are 
the length of the source and target sentences.
For each token $y_{i}$, $(x,y_{1:i-1})$ is called \textit{translation context}.
Let $h_{*}$ be the hidden representations for tokens, then the probability distribution over vocabulary $v$ for token $y_i$, given the translation context, is:
\begin{align}
     p_{\text{MT}}(y_{i}|x,y_{1:i-1}) = \frac{\exp (h_{y_i}^T\cdot h_{i-1})}{\sum_v \exp(h_v^T\cdot h_{i-1})}.
\end{align}
Beam search \cite{bahdanau2014neural,li2016mutual,vijayakumar2016diverse} is normally applied for decoding. 

\paragraph{\textit{k}NN-MT.}
The general idea of $k$NN-MT is to combine the information from $k$ nearest neighbors from a large-scale datastore $S$, when calculating the probability of generating $y_i$.
Specifically, $k$NN-MT first constructs the datastore $\mathcal{S}$ using key-value pairs $(f(x, y_{1:i-1}), y_i)$, where the key is the mapping representation of the translation context 
$h_{i-1}$ for all time steps of all sentences using function $f(\cdot)$, and the value is the  gold target token $y_i$.
The complete datastore is written as $\mathcal{S}=\{(k,v)\}=\{(f(x, y_{1:i-1}), y_i), \forall y_i \in y \}$. 
Then, for each query $q = f(x, y_{1:i-1})$, $k$NN-MT searches through the entire datastore $\mathcal{S}$ to retrieve $k$ nearest translation contexts along with the corresponding target tokens $\mathcal{N}=\{k_j,v_j\}_{j=1}^k$. Last, the retrieved set is transformed to a probability distribution by normalizing and aggregating the negative $\ell_2$ distances, $-d(\cdot, \cdot)$, using the softmax operator with temperature $T$. $p_\text{kNN}(y_i|x, y_{1:i-1})$ can be expressed as follows:
\begin{align}
&  p_{k\text{NN}}(y_i|x, y_{1:i-1})\nonumber\\
&=\frac{\sum_{(k_j,v_j)\in\mathcal{N}}\mathds{1}_{y_i=v_j}\left\{\exp(-d(q, k_j)/T)\right\}}{Z}, \label{knn} \\
& Z=\sum_{(k_j,v_j)\in\mathcal{N}}\exp(-d(q, k_j)/T)\nonumber 
\end{align}

The final probability for the next token in $k$NN-MT, $p(y_{i}|x,y_{1:i-1})$, is a linear interpolation of 
$p_{\text{MT}}(y_{i}|x,y_{1:i-1})$ and $p_{k\text{NN}}(y_{i}|x,y_{1:i-1})$ with a tunable hyper-parameter $\lambda$:
\begin{align}
 p(y_{i}|x,y_{1:i-1})= & \lambda p_{k\text{NN}}(y_{i}|x,y_{1:i-1})+\nonumber\\& (1-\lambda)p_{\text{MT}}(y_{i}|x,y_{1:i-1})
 \label{equation:final_score}
\end{align}

The problem for $k$NN-MT is at each decoding step, a beam search with size $B$ needs to perform $B\times k$ times nearest neighbor searches on the full datastore $S$. 
It is extremely time-intensive 
 when the datastore size $S$ or the beam size is large \citep{khandelwal2020nearest}.

\subsection{Fast $k$NN-MT}
\label{sec:fast_knn_mt}

\begin{figure*}[htb]
     \center{\includegraphics[scale=0.43]{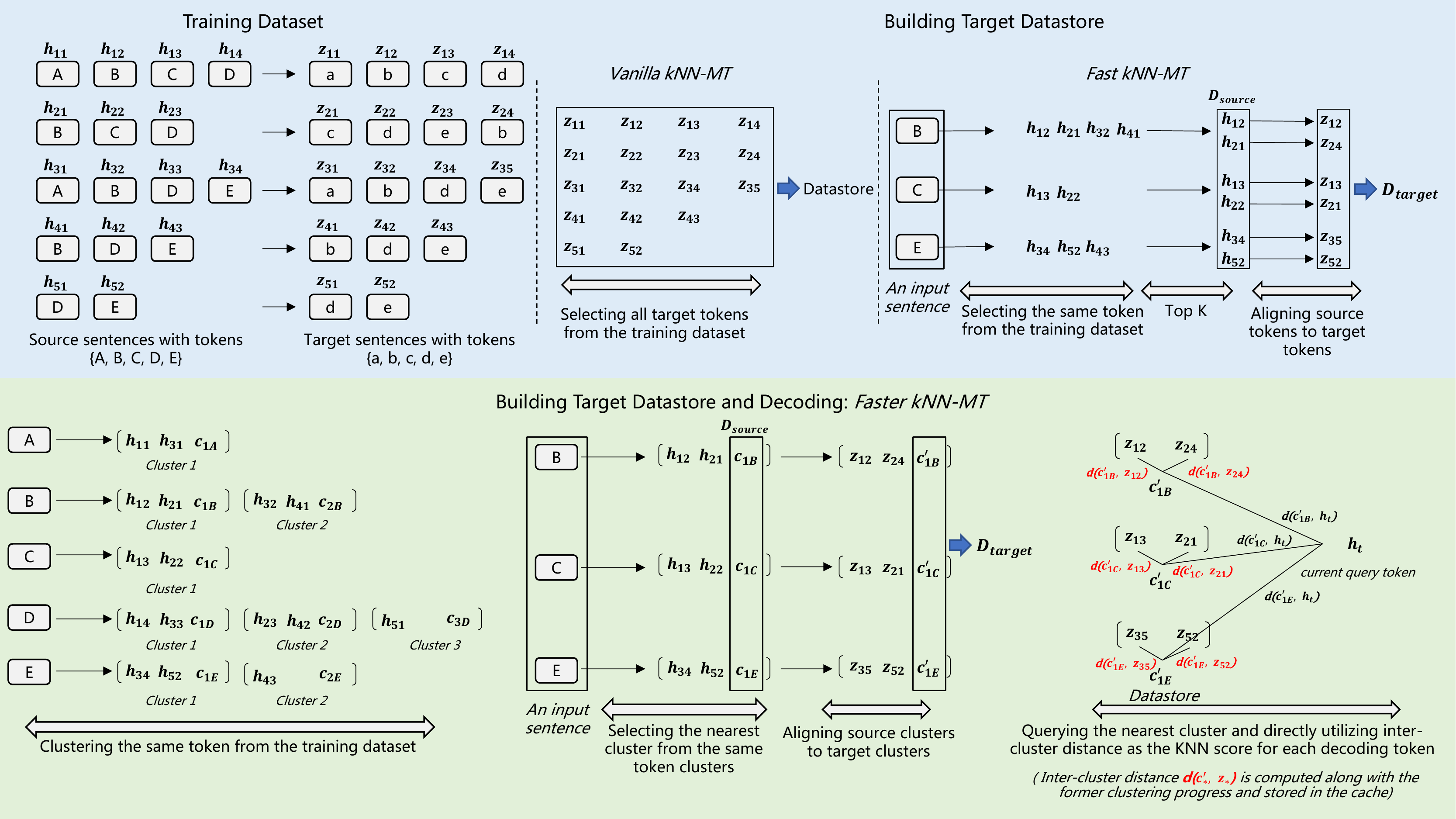}}
    \caption{Comparison between vanilla $k$NN-MT, Fast $k$NN-MT and our proposed Faster $k$NN-MT. For our proposed Faster $k$NN-MT (bottom, green), there are three core steps. (1) {\it Clustering} (bottom, left): We cluster all occurrences of a particular token type from training set into $g$ different groups. (2) {\it Datastore construction} (bottom, middle): Given a test example containing three tokens $\{B,C,E\}$, we first choose the nearest cluster for each source token. Then the selected clusters are aligned to their target clusters. The concatenation of all the centroids in the aligned target clusters constitutes the datastore for the current input. (3) {\it Decoding} (bottom, right): At each decoding step, we query the nearest cluster and directly use inter-cluster distances which is computed along with the former clustering progress and stored in the cache as the $k$NN score for each decoding token.}
    \label{fig:progress}
\end{figure*}

To alleviate time complexity issue in $k$NN-MT, \citet{meng2021fast} proposed Fast $k$NN-MT, which constructs a significantly smaller datastore for the nearest neighbors.
Fast $k$NN-MT consists of the following three steps (also illustrated on the right side of blue part in \Cref{fig:progress}).



\paragraph{Building a Smaller Source Side Datastore.} 
For each source token in the test example, Fast $k$NN-MT limits the $k$NN search to 
tokens of the same token type, in contrast to the whole corpus as in vanilla $k$NN-MT.
Specifically, for each source token of the current test sentence, Fast $k$NN-MT selects the top 
$c$ nearest neighbors from tokens of the same token type in the the source token corpus, rather than from the whole corpus. 
The datastore on the source side $D_\text{source}$ consists of selected nearest neighbors of all constituent tokens within the source sentence.



\paragraph{Transforming Source Datastore to Target Datastore.}
As $k$NN-MT collects the $k$ nearest target tokens during inference, $D_{\text{source}}$ needs to be transformed to a datastore on the target side. \citet{meng2021fast} leverages the FastAlign toolkit \citep{dyer-etal-2013-simple} to link each source token in $D_\text{source}$ to its correspondence on the target side, forming $D_\text{target}$. Each instance in $D_\text{target}$ is a tuple consisting of the aligned target token mapped from the source token and its high-dimensional representation.

\paragraph{Decoding.} At each time step $t$, the representation $h_{t-1}$ produced by the decoder is used to query the target side representations in $D_\text{target}$ 
to search the
 $k$ nearest target neighbors. Then the $k$NN-based decoding probability $p_{k\text{NN}}$ is computed according to the selected nearest neighbors. 
Since $D_\text{target}$ is significantly smaller than the corpus as a datastore, which is used in $k$NN-MT,  
Fast $k$NN-MT is orders of magnitude faster than vanilla $k$NN-MT.

\section{Our Proposed Method: Faster $k$NN-MT}
\label{sec:our_method}

We observe two key issues that hinders the running time efficiency in Fast $k$NN-MT:
(1)
To construct $D_\text{source}$, we need to go through all tokens
in the training set of the same token type.
It can still be time-consuming when the query token is a high-frequency word (e.g., ``is'', ``the'').
(2) The size of $D_\text{target}$ can be large, as 
$D_\text{source}$ combines $c$ nearest neighbors of all input tokens, making it proportional to the size of the source input. 

In this work, we propose Faster $k$NN-MT to tackle these issues. The core idea of our method is to enable a much faster $k$NN search through a hierarchical strategy. 
Faster $k$NN-MT first group tokens of the same type into clusters (in \Cref{sec:clustering_on_the_source_side}). Then, the distance between the query and a data point in the datastore is estimated by (1) the distance between the query and the centroid of the cluster that a data point belongs to, and (2) the distance between the data point and the cluster centroid (in \Cref{sec:clustering_on_the_target_side}). We provide an overview of our proposed method in \Cref{fig:progress} and use an example (in \Cref{sec:example}) to demonstrate our method.


\subsection{Obtaining  $D_{\text{target}}$ on the Target Side}
\label{sec:clustering_on_the_source_side}

Our first step is to construct a datastore on the source side. 
For each token type, we cluster all tokens in the training set of that token type into $g$ different clusters.
Clusters are obtained by using $k$-means clustering algorithm on the token representations, which are the last layer representations from  a pretrained MT model as in vanilla $k$NN-MT.
$g$ is the hyperparameter. 
At test time, for a given source token, we make an approximating assumption that its nearest neighbors should all come from its nearest cluster. Experimental results show that this approximation works well. 
In this work, the nearest clusters are identified based on the $\ell_2$ distance between the representation of the source token and the cluster centroid. 
We combine all selected nearest clusters of all constituent tokens of the source input to constitute the cluster-store on the source side, denoted by $D_{\text{source}}^{\text{cluster}}$.

We then construct the datastore on the target side, as the source datastore can not be readily used to search for nearest neighbors of target tokens during decoding.
We directly map selected source clusters to their corresponding target clusters, since the target correspondence for each token in each source cluster can be readily obtained using FastAlign \citep{dyer-etal-2013-simple}.
The target cluster corresponding to a source cluster is the union of all  target tokens  corresponding to source tokens in that source cluster.
We denote the cluster-store on the target side as $D_{\text{target}}^{\text{cluster}}$. 
The concatenation of constituent data points in clusters within $D_{\text{target}}^{\text{cluster}}$ constitute the target datastore, denoted by $D_{\text{target}}$.
In practice, the mapping between source and target clusters can be obtained in advance and cached.

\subsection{Selecting $k$NN on the Target Side}
\label{sec:clustering_on_the_target_side}

We now have the target datastore, the next step is to run nearest neighbor search in each decoding step. $k$ nearest neighbors of $h_{t-1}$ from $D_{\text{target}}$ is selected by ranking $d(h_{i-1}, z_j)$, the distance between the query representation $h_{i-1}$ and a point $z_j$ in the target datastore. To simplify notations and without loss of generality, below we will only consider a 1 nearest neighbor situation, we note $k$ nearest neighbors can be computed in a similar way. 

We obtain the index for the nearest data point by:
\begin{align}
 \text{index for 1 NN} = \argmin_j d(h_{i-1}, z_j).
 \label{eq2}
\end{align}
The key point of Faster $k$NN-MT is to approximately compute the distance $d(h_{i-1}, z_j)$ by decoupling it into two parts: (1) $d(c_l, h_{i-1})$, which is the distance between the $h_{i-1}$ and the cluster centriod $c_l$ that a given target point $z_j$ belongs to; and (2) $d(c_l, z_j)$, which is the distance between the cluster centriod and the point $z_j$:
\begin{align}
d(h_{i-1}, z_j) \approx  d(c_l, h_{i-1}) + d(c_l, z_j).
\label{eq:decouple}
\end{align}

In this work, to enable faster computations, we approximate the minimum of the addition of $d(c_l, h_{i-1})$ and $d(c_l, z_j)$ in \Cref{eq:decouple} by finding the minimum for each term: (1) finding the nearest cluster by $ l = \argmin_l d(c_l, h_{i-1})$, and (2) finding the nearest neighbor by $
 j = \argmin_j d(c_l, z_j)$.
This approximation works well because clusters in $D_{\text{target}}^{\text{cluster}}$ are distinct: recall when we construct $D_{\text{source}}^{\text{cluster}}$, for each source token, we find its nearest cluster from clusters of the same token type, and add the cluster to $D_{\text{source}}^{\text{cluster}}$. 
Each cluster in $D_{\text{source}}^{\text{cluster}}$ corresponds to a unique token type.
As clusters in  $D_{\text{source}}^{\text{cluster}}$ are mapped to  $D_{\text{target}}^{\text{cluster}}$ in one-to-one correspondence, 
clusters in  $D_{\text{target}}^{\text{cluster}}$ should correspond to different token types, and are thus different.

 
We observe our two-step procedure for finding the minimum data point above is extremely computationally effective. We only need to go over $O(n)$ clusters for finding the nearest cluster. Ranks of data points based on distances to cluster centroid can be computed in advance and cached, meaning no computations required at the test time.



\subsection{An Illustrative Example}
\label{sec:example}

In this section, we work through the example in \Cref{fig:progress} (in green) to better illustrate our procedures. We assume that there are five kinds of source tokens $\{A, B, C, D, E\}$ and five kinds of target tokens $\{a,b,c,d,e\}$ in the training set. We use $h_{*}$ and $z_{*}$ for the representations of each token generated by the last layer of the pre-trained MT model in the source side and target side, respectively.

\paragraph{Obtaining $D_\text{target}$ on the Target Side.}

We first cluster tokens of the same type in the training set into at most $g$ clusters. In this example, we take $g$=3 and then generate clusters for tokens based on their hidden representations. In each cluster, besides the specific tokens, we also calculate the corresponding centroid of that cluster, denoted as $\{c_{\text{type}}\}$. For instance, for token $B$, we generate two clusters $\{h_{12}, h_{21}\}$ and $\{h_{32}, h_{41}\}$, and assign the cluster centroid $c_{1B}$ and $c_{2B}$ to these two clusters.

As we need to build a datastore on the target side for decoding, we now 
construct the cluster-store on the source side $D_{\text{source}}^{\text{cluster}}$, by querying the nearest cluster according to the distance between the representation of a specific token and the cluster centroid representations for this token. Suppose that the cluster $\{h_{12}, h_{21}, c_{1B}\}$ is the nearest cluster for token $B$, among two clusters of $B$. Similarly, we assume the cluster $\{h_{13}, h_{22}, c_{1C}\}$ is the nearest cluster for token $C$ and the cluster $\{h_{34}, h_{52}, c_{1E}\}$ is the nearest cluster for token $E$. The concatenation of all the tokens of above three selected clusters constitute the source side $D_{\text{source}}^{\text{cluster}}$ for the given sentence $\{B,C,E\}$.

To construct the cluster-store on the target side $D_{\text{target}}^{\text{cluster}}$, we use FastAlign toolkit for the constituted $D_{\text{source}}^{\text{cluster}}=\{\{h_{12}, h_{21}\}, \{h_{13}, h_{22}\}, \{h_{34}, h_{52}\}\}$ to find the mapped representation in the target side. Suppose that $\{\{z_{12}, z_{24}\}, \{z_{13}, z_{21}\}, \{z_{35}, z_{52}\}\}$ is the mapped set from $D_{\text{source}}^{\text{cluster}}$. We then associate the centroid for each target cluster $c_{1B}'$, $c_{1C}'$, and $c_{1E}'$ after averaging all the representations of each target cluster.
The target datastore $D_{\text{target}}$ contains all centroids in $D_{\text{target}}^{\text{cluster}}$.

\paragraph{Selecting $k$NN on the Target Side.} At each decoding step $t$, to collect the $k$ nearest neighbors for the current decoding representation $h_{t}$, we first utilize $h_{t}$ to query the nearest target cluster in the target datastore $D_{target}=\{c_{1B}', c_{1C}', c_{1E}'\}$ according to the distance $d(c_{\text{type}}', h_{t}), \text{type} \in\{1B,1C,1E\}$. We suppose that the cluster $1B$ is chosen for the current decoding representation $h_{t}$. Then we select $k$ nearest neighbors in the inter-cluster representations of the target cluster $cluster_{1B}^{target}=\{z_{12},z_{24}\}$ according to the inter-cluster distances $\{d(c_{1B}', z_{12}), d(c_{1B}', z_{24})\}$. All above distances are computed in advance.

\subsection{Comparisons to Fast $k$NN-MT}


We now compare the speed and space complexity of Faster $k$NN-MT against Fast $k$NN-MT. 

Let $g$ be the number of clusters, $c$ be the number of nearest neighbors for NN search in $D_{\text{source}}$, 
$F$ be the frequency of the source token,
$d$ be the representation dimensionality,  
and $n$ be the length of the source sentence in the test example.

\paragraph{Time Complexity.}
For Fast $k$NN-MT,
to construct datastore on the source side, it needs to
search $k$-nearest neighbors from $F$ source tokens on average and 
construct $D_{\text{source}}$ with a size of $cn$ with 
 a time complexity of $O(Fdcn)$. 
For decoding, the size of $D_{\text{target}}$ is the same as 
$D_{\text{source}}$. 
For each decoding step, it needs to search the $k$ nearest neighbors from the datastore with size $cn$, making the time complexity for each decoding step being $O(kdcn)$.
We assume that the length of the decoded target is very similar to the source length, i.e., $n$.
The time complexity for decoding is thus $O(kdcn^2)$.
Summing all, the time complexity for Fast $k$NN-MT is $O(Fdcn+kdcn^2)$.

For Faster $k$NN-MT, to construct $D_{\text{source}}^{\text{cluster}}$, we only need to search $k$-nearest clusters from $g$ source clusters, which  
leads to a time complexity of $O(gdn)$ for a source of length $n$. For each token in the source, we only select the nearest neighbor, which leads to the size of  $D_{\text{source}}^{\text{cluster}}$ being $n$. Due to the one-to-one correspondence between source cluster and target cluster, 
the size of  $D_{\text{target}}^{\text{cluster}}$ is also $n$.
At each decoding step, we search the nearest cluster from
$D_{\text{target}}^{\text{cluster}}$, 
leading a  
time perplexity of  $O(dn)$ for each step,
and thus  $O(dn^2)$ for the whole target. 
 Since all distances and ranks are computed in advance and cached,
  nearest neighbors in the selected cluster are picked with $O(1)$
  time perplexity. 
 The Overall time complexity of Faster $k$NN-MT is $O(gdn+dn^2)$ which is significantly smaller than $O(Fdcn+kdcn^2)$
 of Fast $k$NN-MT.

\paragraph{Space Complexity.} For space complexity, for Fast $k$NN-MT,
the size of $D_{\text{source}}$ and $D_{\text{target}}$ are both $c\times n$, 
leading to a space complexity $O(cnd)$, where $d$ denotes the representation dimensionality;
while for Faster $k$NN-MT, the size of $D_{\text{source}}^{\text{cluster}}$ or 
$D_{\text{target}}^{\text{cluster}}$
is $n$ respectively, leading to a space complexity $O(nd)$. 
This significant saving in space let us 
 increase the batch size with limited GPU memory, which also leads to  a significant speedup.

\section{Experiments}
\label{sec:exp}



\subsection{Datasets}

We experiment with two types of datasets: traditional bilingual and domain adaptation datasets. 
Table \ref{tab:datasets_details} shows the statistics for these datasets.

\paragraph{Bilingual Datasets.} We use  WMT'14 English-French\footnote{\url{http://www.statmt.org/wmt19/translation-task.html}} and WMT'19 German-English.\footnote{\url{http://www.statmt.org/wmt14/translation-task.html}}.  
We follow protocols in \citet{ng2019facebook}, including 
applying language identification filtering and only keep sentence pairs with correct language on both source and target side; 
 removing sentences longer than 250 tokens as well as sentence pairs with a source/target length ratio exceeding 1.5; 
 normalizing punctuation and tokenize all data with the Moses tokenizer \citep{koehn2007moses};
and utilizing subword segmentation \citep{sennrich2016neural} doing joint byte pair encodings (BPE) with 32K split operations for WMT'19 German-English and 40K split operations for WMT'14 English-French. 


\paragraph{Domain Adaptation Datasets.} 
We use Medical, IT, Koran and Subtitles domains in the domain-adaptation benchmark \citep{koehn2017challenges}. For each domain dataset, we split it into train/dev/test sets and clean these sets following protocols in  \citep{aharoni2020unsupervised}. 

\begin{table*}[h!]
    \centering
    \resizebox{.8\textwidth}{!}{
    \begin{tabular}{l|cc|cccc}
    \toprule
         & \multicolumn{2}{c|}{\textit{Bilingual Translation}} & \multicolumn{4}{c}{\textit{Domain Adaptation}} \\
         & \textbf{WMT'14 En-Fr} & \textbf{WMT'19 Ge-En} & \textbf{Medical} & \textbf{IT} & \textbf{Koran} & \textbf{Subtitles} \\\midrule
        Sentence pairs & 35M & 32M & 0.25M & 0.22M & 0.02M & 0.5 \\
        Maximum source sentence length & 250 & 250 & 469 & 704 & 252 & 65 \\
        Average source sentence length & 31.8 & 27.9 & 13.9 & 9.0 & 19.7 & 7.6 \\
        Number of tokens & 1.1G & 0.9G & 3.5M & 2.0M & 0.3M & 3.9M \\
        Number of token types & 44K & 42K & 18K & 21K & 7K & 23K \\
        Maximum token frequency & 62M & 40M & 0.18M & 0.11M & 0.03M & 0.4M\\
        Average token frequency & 26K & 23.8K & 374 & 182 & 74 & 237 \\\bottomrule
    \end{tabular}
    }
    \caption{Dataset statistics for  bilingual translation datasets and domain adaptation datasets.}
    \label{tab:datasets_details}
\end{table*}

\subsection{Implementation Details}


\paragraph{Base MT Model.} We directly use the Transformer based model provided by the FairSeq \citep{ott2019fairseq} library as the vanilla MT model.\footnote{\url{https://github.com/pytorch/fairseq/tree/master/examples/translation}}
Both the encoder and the decoder have 6 layers. We set the dimension of word representations to 1,024, the number of multi-attention heads to 6 and the inner dimension of feedforward layers to 8,192. 

\paragraph{Quantization.}

To make sure all the token representations can be loaded into memory, we perform the product quantization \citep{jegou2010product}. 
For each token representation $x\in \mathbb{R}^{D}$, we first split it into $M$ subvectors: $[x_{1},x_{2},...,x_{M}]$ with the same dimension $d=D/M$.
We then train the product quantizer using the following objective function:
\begin{align}
\min_{q^{1}, ..., q^{M}}\sum_{x}\sum^{M}_{m=1}\parallel x_{m}-q_{m}(x_{m})\parallel^{2},
\end{align}
where ${q_{i}\, (1\leq i\leq M)}$ denotes $M$ sub-quantizers used to map a subvector $x_{m}\in \mathbb{R}^{d}$ to a codeword in a subcodebook $C_{m}$.
Lastly, we leverage the $M$ quantizers $q_{1},...,q_{M}$ to compress the high dimensional vector $x$ to $M$ codewords. We set $M$ to be 128 in this work.

\paragraph{FAISS $k$NN Search.} 
We use FAISS \citep{johnson2019billion}, a toolkit for  approximate nearest neighbor search, to speed up the process of KNN search.
FAISS firstly samples $N$ data points from the full dataset,
which are 
clustered into $M$ clusters. The remaining data in the  dataset are then mapped to these $M$ clusters.
For a given query, it first queries the nearest cluster and does brute force search within this cluster. In this paper, we directly adopt the brute force search for tokens with frequency lower than 30,000; 
For tokens with frequency larger than 30,000, 
we do the search using FAISS toolkit for tokens.

\paragraph{Other Details} We use the $\ell_{2}$ distance to compute the similarity function in $k$-means clustering and use FAISS \citep{8733051} to cluster all reference tokens on the source side. 
The number of clusters for each source token type is set to $f/m$, where $f$ is the frequency of  the type token and $m$ is the hyper-parameter controlling the number of clusters,
which is set to 2,048. 




\subsection{Results on Bilingual Datasets}
\label{sec:ablation_experiments}

To tangibly understand the behavior of each module of Faster $k$NN, we conduct ablation experiments on the two WMT datasets by combining each module of Faster $k$NN respectively with Fast $k$NN. We experiment with the following two setups:
\begin{itemize}
\item \label{item_source_cluster} \textbf{Fast $k$NN with
Faster $k$NN's Source datastore}: We replace the source-side datastore of Fast $k$NN-MT
with the datastore $D_{\text{source}}^{\text{cluster}}$ 
from Faster  $k$NN-MT.
This is to test the individual influence of clustering tokens 
of the same token type when constructing source side datastore,
as opposed to using all tokens of the same token type as the datastore in Fast $k$NN. 
 More specifically,  we first 
 construct 
 the cluster-store on the source side $D_{\text{source}}^{\text{cluster}}$ as in Faster $k$NN-MT.
 Then, 
 we conduct the token-level mapping to map the source side cluster-store $D_{\text{source}}^{\text{cluster}}$ to target side datastore $D_{\text{target}}$.  
 $D_{\text{target}}$  is then integrated to Fast $k$NN-MT as the target datastore for each decoding step. 
\item \textbf{Faster $k$NN without Cached Inter-cluster distance:}
At each decoding step, we obtain the top-$k$ nearest neighbors of a target query by 
 directly computing the distance the query with data points in 
$D_{\text{target}}^{\text{cluster}}$, instead of using cached inter-cluster distance for speed-up purposes. 
This is to test whether the inter-cluster distance approximation in \Cref{eq:decouple} 
for selecting top-k nearest neighbors 
results in a performance loss. 
Specifically, 
as in Faster $k$NN, 
we use the target side cluster-store $D_{\text{target}}^{\text{cluster}}$ mapped from $D_{\text{source}}^{\text{cluster}}$  as the target side datastore. 
At each decoding step,
instead of 
selecting the top-k points based on inter-cluster distances as in Faster $k$NN, we select the top-1 nearest cluster from $D_{\text{target}}^{\text{cluster}}$ and chose top-$k$ nearest target token representations by directly computing the distance between $h_{t}$ and data points. 
\end{itemize}



\paragraph{Main Results.} We report the SacreBLEU scores \citep{post-2018-call} in Table \ref{tab:bleu_result_on_wmt}.
We observe that our proposed Faster $k$NN-MT model achieves comparable BLEU scores to vanilla $k$NN-MT and Fast $k$NN-MT on English-French and German-English datasets, but with a significant speedup.

\begin{figure}[h!]
     \center{\includegraphics[scale=0.22]{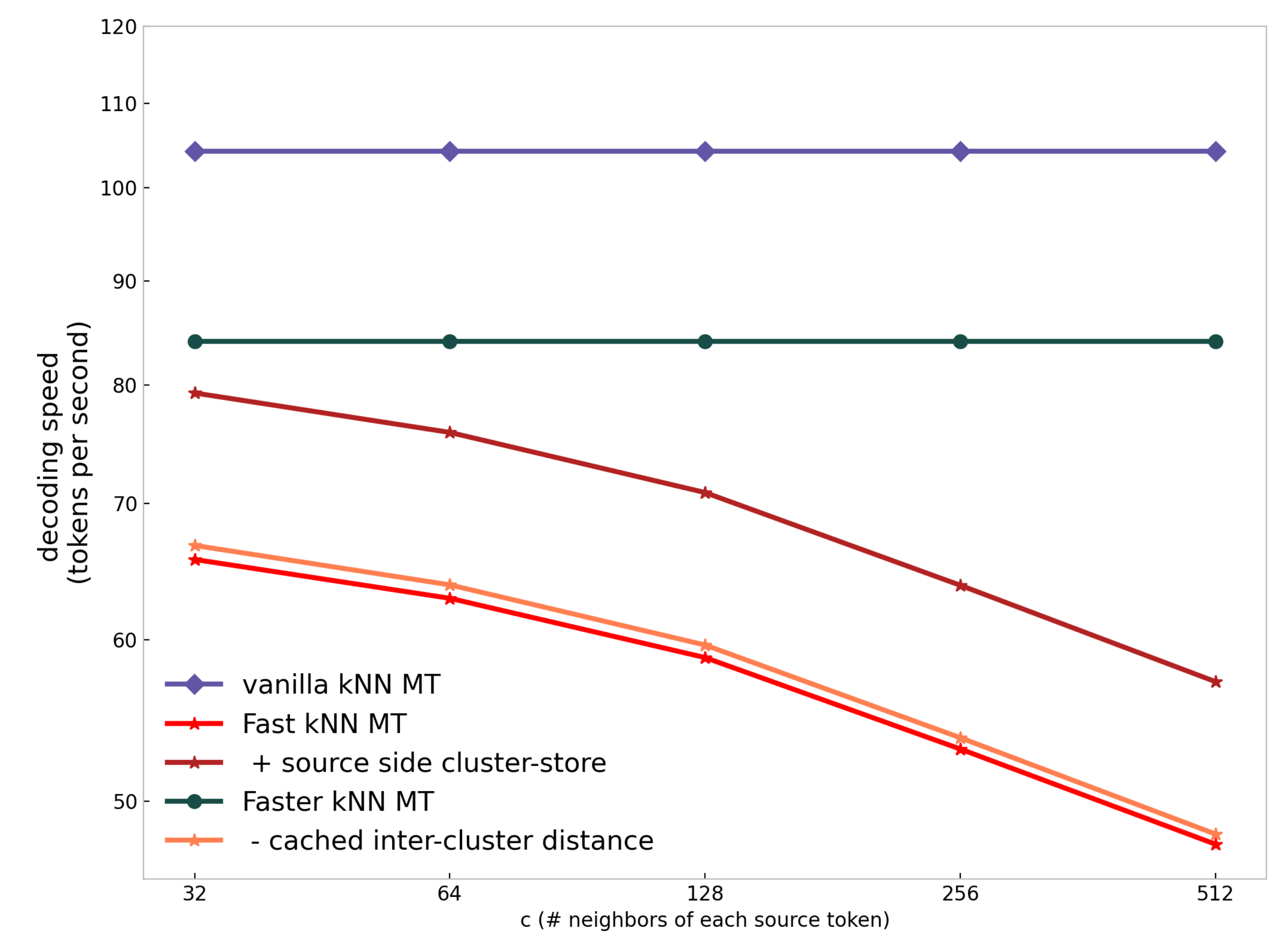}}
    \caption{Speed comparison between Base MT, Fast $k$NN-MT, Faster $k$NN-MT and two ablation strategies.}
    \label{fig:speed}
\end{figure}

In \Cref{fig:speed}, we show the speed comparison between vanilla MT, Fast $k$NN-MT and Faster $k$NN-MT.
Results for vanilla $k$NN-MT are just omitted as it is two orders of magnitude slower than vanilla MT \cite{khandelwal2020nearest,meng2021fast}. For Fast $k$NN-MT we observe that the speed advantage gradually diminishes as the number of nearest neighbors
in $D_{\text{source}}$ increases. For Faster $k$NN-MT, since the size of datastore $D_\text{target}$ used at each decoding step is fixed to the length of source test sentence, it would not suffer speed diminishing when the length of the input source get greater.

\paragraph{Fast $k$NN-MT with Faster $k$NN’s source side cluster-store $D_{\text{source}}^{\text{cluster}}$.} To build datastore $D_{source}$, for each source token in the test example, Fast $k$NN-MT selects the top $c$ nearest neighbors from tokens of the same token type in the source token corpus. Note that not all the $c$ nearest neighbors can be clustered in the same one cluster in $D_{\text{source}}^{\text{cluster}}$, and that Faster $k$NN’s only picks one cluster on the source side. The results for that setup is thus lower than Fast $k$NN-MT. For speed comparison between this setup and Fast $k$NN-MT shown in figure \ref{fig:speed}, since the size of $D_{target}$ used at each decoding step 
for the two setups
is approximately equal, the time consumption is almost equal.

\paragraph{Faster $k$NN-MT without cached inter-cluster distance.} The BLEU scores on WMT German-English and WMT English-French datasets is 
comparable
between the proposed setup,Fast $k$NN-MT and Faster $k$NN-MT. For speed comparison shown in figure \ref{fig:speed}, we can see that the speed of the current setup is faster than Fast $k$NN-MT but still slower than Faster $k$NN-MT, especially as  the number of nearest neighbors queried at each decoding step increases. This result shows that the inter-cluster distance approximation in Eq.\ref{eq:decouple} does improve the speed of Faster $k$NN-MT at each decoding step, while the performance loss is not significant.

\begin{table}[h!]
    \centering
    \resizebox{.49\textwidth}{!}{
    \begin{tabular}{lll}\toprule
        \textbf{Model} & \textbf{De-En} & \textbf{En-Fr}  \\\midrule
        Base MT & $37.6$ & $41.1$ \\
        + $k$NN-MT & ${39.1}_{(+1.5)}$ & ${41.8}_{(+0.7)}$ \\
        + Fast $k$NN-MT & ${39.3}_{(+1.7)}$ & ${41.7}_{(+0.6)}$ \\\midrule
        Faster $k$NN-MT  & ${39.3}_{(+1.7)}$ & ${41.6}_{(+0.5)}$ \\\midrule
        \multicolumn{3}{l}{\textit{Ablation Experiments}}\\
        Fast $k$NN-MT + $D_{\text{source}}^{\text{cluster}}$ & ${39.1}_{(+1.5)}$ & ${41.4}_{(+0.3)}$ \\
        Faster $k$NN-MT - cached inter-cluster dist. & ${39.5}_{(+1.9)}$ & ${41.6}_{(+0.5)}$ \\\bottomrule
    \end{tabular}
    }
     \caption{SacreBLEU scores on WMT'14 En-Fr and WMT'19 Ge-En datasets.}
    \label{tab:bleu_result_on_wmt}
\end{table}

\begin{table*}[h!]
    \centering
    \resizebox{.8\textwidth}{!}{
    \begin{tabular}{llllll}\toprule
        \textbf{Model} & \textbf{Medical} & \textbf{IT} & \textbf{Koran} & \textbf{Subtitles} & \textbf{Average}  \\\midrule
        \citet{aharoni2020unsupervised} & $54.8$ & $43.5$ & $21.8$ & $27.4$ & $47.2$ \\\hline
        Base MT & $39.9$ & $38.0$ & $16.3$ & $29.2$ & $30.9$ \\
        + $k$NN-MT & ${54.4}_{(+14.5)}$ & ${45.8}_{(+7.8)}$ & ${19.4}_{(+3.1)}$ & ${31.7}_{(+2.5)}$ & ${37.8}_{(+6.9)}$ \\
        + Fast $k$NN-MT & ${53.6}_{(+13.7)}$ & ${45.5}_{(+7.5)}$ & ${21.2}_{(+4.9)}$ & ${30.5}_{(+1.3)}$ & ${37.7}_{(+6.8)}$  \\
        + Faster $k$NN-MT  & ${52.7}_{(+12.8)}$ & ${44.9}_{(+6.9)}$ & ${20.4}_{(+4.1)}$ & ${30.2}_{(+1.0)}$ & ${37.1}_{(+6.2)}$ \\\bottomrule
    \end{tabular}
    }
     \caption{SacreBLEU scores on four domain datasets: Medical, IT, Koran and Subtitles.}
    \label{tab:bleu_result_on_domain}
\end{table*}

\subsection{Results on Domain Adaptation Datasets}

For domain adaptation, we evaluate the base MT model and construct datastore within four German-English domain parallel datasets: Medical, IT, Koran and Subtitles, which are originally provided in \citep{koehn2017challenges}. Results are shown in Table \ref{tab:bleu_result_on_domain}. 
We observe that our proposed Faster $k$NN-MT model achieves comparable BLEU scores to Fast $k$NN-MT and vanilla $k$NN-MT on the four datasets of domain adaption task, and similar to the performance on the two WMT datasets the time and space consumption of Faster $k$NN-MT are both much smaller than Fast $k$NN-MT and vanilla $k$NN-MT.





\section{Related Work}
\label{sec:related_work}

\paragraph{Neural Machine Translation.}
Recent advances on neural machine translation are build upon encoder-decoder architecture \citep{sutskever2014sequence, cho2014learning}. The encoder infers a continuous representation of the source sentence, while the decoder is a neural language model conditioned on the encoder output. The parameters of both models are learned jointly to maximize the likelihood of the target sentences given the corresponding source sentences from a parallel corpus. 
More robust and expressive neural MT systems have also been developed \citep{guo2020incorporating, zhu2020incorporating, kasai2021deep, kasai2021finetuning, lioutas2020time, peng2021random, tay2021synthesizer, li2020sac, liu2020understanding, nguyen2019transformers, wang2019learning, xiong2020layer} based on attention mechanism \citep{bahdanau2014neural, luong2015effective}.

\paragraph{Retrieval Augmented Model.} 

Retrieval augmented models additionally use the input to retrieve a set of relevant information, compared to standard neural models that directly pass the input to the generator.
 Prior works have shown the effectiveness of retrieval augmented models in improving the performance of a variety of natural language processing tasks, including language modeling \citep{khandelwal2019generalization,meng2021gnn}, question answering \citep{guu2020realm, lewis2020pre, lewis2020retrieval, xiong2020approximate},
 text classification \cite{lin2021bertgcn},
 and dialog generation \citep{fan2020augmenting, thulke2021efficient, weston2018retrieve}. 

For neural MT systems, \citet{zhang2018guiding} retrieves target $n$-grams to up-weight the reference probabilities. 
\citet{bapna2019non} attend over neighbors similar to $n$-grams in the source using gated attention \citep{cao-xiong-2018-encoding}.
\citet{tu2017learning} made a difference saving the former translation histories with the help of cache-based models \citep{grave2016improving}, and the model thus can deal with a changing translation contexts.

There are also approaches improving the translation results by directly retrieving the example sentence in the training set. At the beginning of the machine translation, a lot of techniques focus on translating sentences by analogy \citep{Nagao1981AFO}. These techniques identify the similar examples based on edit distance \citep{doi2005example} and trigram contexts \citep{van2007memory}. For recently, \citet{gu2018search} collected sentence pairs according to the given source sentence from the small subset of sentence pairs from the training set leveraging an off-the-shelf search engine. Since these techniques focus on sentence-level machining, they will be hard to handle facing large and changing contexts.
To take more advantage of neural context representations, \cite{khandelwal2020nearest} proposed $k$NN-MT that it simply collects all the target representations in the training set, and constructs a much larger datastore than the above approaches. However, the approaches described above mainly focus on either efficiency or performance. To benefit from retrieval augmented model without loss of efficiency, \citet{meng2021fast} proposed the Fast $k$NN-MT. This work offers a further speed-up than Fast $k$NN-MT.


\section{Conclusion}
\label{sec:conclusion}

In this paper, we propose Faster $k$NN-MT, a method to further speed up the previous Fast $k$NN-MT model. Our method improves the speed to only 1.2 times slower than base MT, compared to Fast $k$NN-MT which is 2 times slower.
The core idea of Faster $k$NN-MT is to constrain the search space when constructing the datastore on both source side and target side. We leverages $k$-means clustering for only querying the centroid of each cluster instead of all examples from the datastore. Experiments demonstrate that this strategy is more efficient than Fast $k$NN-MT with minimal performance degradation.

\bibliography{custom}
\bibliographystyle{acl_natbib}

\appendix


\end{document}